\documentclass[manuscript, screen]{acmart}
\usepackage{bbold}
\setcopyright{acmlicensed}
\copyrightyear{2024}
\acmYear{2024}
\makeatletter
\let\@authorsaddresses\@empty
\makeatother

\title{Segment Discovery: Enhancing E-commerce Targeting}
\author{Qiqi Li}
\authornote{Also with Univeristy of California, Santa Barbara}
\affiliation{%
   \institution{Amazon}
   \country{USA}
}
\author{Roopali Singh}
\affiliation{%
   \institution{Amazon}
   \country{USA}
   }
\author{Charin Polpanumas}
\affiliation{%
   \institution{Amazon}
   \country{Japan}
   }
\author{Tanner Fiez}
\affiliation{%
   \institution{Amazon}
   \country{USA}
   }
\author{Namita Kumar}
\affiliation{%
   \institution{Amazon}
   \country{USA}
   }
\author{Shreya Chakrabarti}
\affiliation{%
   \institution{Amazon}
   \country{USA}
   }

\acmConference[RecSys’24 CONSEQUENCES Workshop]{Recommender Systems Conference, CONSEQUENCES Workshop.}{October 14-18,
  2024}{Bari, Italy}

\begin{document}

\begin{abstract}
Modern e-commerce services frequently target customers with incentives or interventions to engage them in their products such as games, shopping, video streaming, etc. This customer engagement increases acquisition of more customers and retention of existing ones, leading to more business for the company while improving customer experience. Often, customers are either randomly targeted or targeted based on the propensity of desirable behavior. However, such policies can be suboptimal as they do not target the set of customers who would benefit the most from the intervention and they may also not take account of any constraints. In this paper, we propose a policy framework based on uplift modeling and constrained optimization that identifies customers to target for a use-case specific intervention so as to maximize the value to the business, while taking account of any given constraints. We demonstrate improvement over state-of-the-art targeting approaches using two large-scale experimental studies and a production implementation.
\end{abstract}

\maketitle

\section{Introduction}
Promotions and discounts have become key components of modern e-commerce services. Popular promotions include discounts, bundled offers, free services, etc. By offering these promotions, companies aim to increase revenue and customer base, while also improving customer experience. However, such promotions usually incur a cost and can become unsustainable without any guardrails in place. A popular approach is to target customers with high or low propensity for desired behavior. For example, a retail company is likely to target customers who are at risk of leaving if they want to retain its customers by offering certain incentives. 
However, previous studies show that this strategy is ineffective and could be detrimental towards the company objectives \cite{du2019improve, ascarza2018retention, devriendt2021you}. Moreover, additional analysis needs to be done for the choice of propensity score threshold for targeting (e.g., target anyone whose propensity to leave is higher than 0.8), because the wrong threshold may lead to sub-optimal outcomes \cite{ascarza2018retention}.

Each customer responds differently to the same promotion. This motivates the development of our personalized targeting approach that showcases these promotions to a set of customers that will have a positive response due to the treatment offered, while controlling for any business constraints. The contributions of this work are as follows: 1. We present a structured two-step methodology for customizing customer targeting strategies in the e-commerce industry. Our framework considers specific business constraints during the allocation of treatments to customers, aiming to optimize desired business outcomes. 2. We illustrate the practical application of our framework by devising targeting strategies for three distinct business scenarios, each characterized by unique outcome objectives and operational constraints. 3. We demonstrate the superiority of our approach over commonly used targeting methods based on propensity thresholds in e-commerce settings across three key business applications using offline policy estimation techniques. 4. We validate the efficacy of our proposed targeting policies against the current baseline approach using a large-scale online A/B test.

\subsubsection*{\textbf{Related Work}} Previously, a similar approach has been used in context of rehabilitation program design \cite{yadav2021optimal} while some companies have directly incorporated cost into the outcome variable of the causal uplift models \cite{zhao2019uplift} and performed constrained optimization for a specific business use-case \cite{goldenberg2020free, albert2022commerce}. However, our approach offers a more generalized framework that is applicable to a wide variety of customer targeting problems.

\section{Methodology}\label{methodology}

The goal is to identify a set of customers to target with a treatment, while optimizing for a metric taking account of any constraints. Consider a set of customers $\mathcal{C} = \{1,2,\ldots, N\}$ and a set of treatments $\mathcal{T} = \{t_0,t_1,\ldots,t_K\}$ to be applied to this customer set. Suppose we have some experimental data, where we observe $\{X_i, T_i, Y_i\}$ for each customer $i = 1,2, \ldots, N$. Here, $X_i$ is a set of covariates associated with customer $i$ and $Y_i$ is the observed outcome of interest when customer $i$ receives treatment $T_i \in \mathcal{T}$. We aim to find an optimal targeting policy $\uppi$, that assigns a treatment to each customer given certain attributes $X$, while optimizing for $Y$. The proposed policy is evaluated using metrics discussed in Section~\ref{eval_metrics} and specific business metrics for each use case.

\subsection{Solution Framework}\label{solution_frame}
Given the problem setup, the goal is to find a set of customers to target with a specific treatment such that it optimizes the outcome, given certain constraints. We propose a two-stage approach: (1) estimate the impact of a treatment on the outcome for each customer, (2) find an optimal set of customers such that the estimated impact is maximized, given a constraint.

\subsubsection*{Stage 1: Uplift Model}
Given the potential outcomes framework \cite{rubincausal2005}, we define $Y_i(t_k)$ as the potential outcome corresponding to customer $i$, if they were to receive treatment $t_k \in \mathcal{T}$. We are interested in estimating conditional
average treatment effect (CATE) of treatment $t_k$. Assuming $t_0$ as the control treatment level, CATE of treatment $t_k$ for an individual $i$ can be defined as $$\tau_k(x_i)=\text{E}[Y_i(t_k) \mid X_i = x_i] - \text{E}[Y_i(t_0) \mid X_i = x_i].$$
A key assumption for identifying CATE is unconfoundedness (strong ignorability) which
means that conditional on covariates, the potential outcomes are independent of the treatment assignment. When it holds, current literature offers a wide range of models for causal estimation such as meta-learners (e.g., S, T and X-learners) and forest-based estimators (e.g., forest doubly robust Learner\cite{oprescu2019orthogonal}, causal forest \cite{wager2018estimation} and causal forest double machine learning estimator \cite{athey2019generalized}). 

\subsubsection*{Stage 2: Constrained Optimization}
Next step is to find an optimal targeting policy, $\uppi$, which assigns treatments to customers such that it maximizes the uplift measured in stage 1. A policy $\uppi$ can be defined as a matrix of order $N \times K$ as $\uppi = [\pi_{ik}]_{N \times K+1}$, where 
$$\pi_{ik}=
\begin{cases}
1, \text{ if treatment } t_k \text{ is assigned to customer } i \\
0, \text{ if treatment } t_k \text{ is not assigned to customer } i
\end{cases}$$
such that $\sum_k \pi_{ik} =1$. The optimal policy $\uppi$ can be found such that,
$$\text{max}_\uppi \sum_{i=1}^{N}\sum_{k=1}^{K} \pi_{ik} w_i \hat{\tau}_k(x_i), \text{ s.t. } \pi_{ik} \in \{0,1\} \ ,\ \sum_k \pi_{ik} =1 \ , \ g(\uppi)\leq c, $$ 
where $\hat{\tau}_k(x_i)$ is the estimate of $\tau_k(x_i)$, $w_i$ is a predefined weight for each customer, $g(.)$ is a convex function of the policy $\uppi$ and $c$ is a predefined constant. For example, there can be budget constraint on the number of customers assigned to each treatment i.e., $g_k(\uppi) = \sum_i \pi_{ik} \leq c_k$ for a treatment $k$.

\subsection{Evaluation Metrics}\label{eval_metrics}
\subsubsection*{Uplift Model Evaluation}
A common evaluation metric for uplift models is the area under the cumulative uplift curve \cite{gutierrez2017causal, rzepakowski2012decision}. We rank the customers by their predicted uplift from the corresponding causal model. The cumulative uplift for a customer ranked $r^{th}$ by the predicted uplift, can be defined as:
$$\text{Cumulative Uplift}_r = (\displaystyle\frac{\sum_{i=1}^r Y_i \mathbb{1}\{\pi_{i0} = 0 \text{ \& } T_i \neq t_0\}}{\sum_{i=1}^r \mathbb{1}\{\pi_{i0} = 0 \text{ \& } T_i \neq t_0\}} - \displaystyle\frac{\sum_{i=1}^r Y_i \mathbb{1}\{\pi_{i0} = 1 \text{ \& } T_i = t_0 \}}{\sum_{i=1}^r \mathbb{1}\{\pi_{i0} = 1 \text{ \& } T_i = t_0 \}}) * \frac{r}{|\mathcal{C}|},$$

where $Y_i$ is the observed outcome, $t_0$ is the control treatment and $\pi_{i0} = 1$, if customer $i$ is assigned treatment $t_0$ by policy $\uppi$ else 0. We can compute the area under the curve (AUC), where the larger AUC value indicates a better model.

\subsubsection*{Offline Policy Evaluation}
The observed data comes from randomized experiments (logging policy) that maps customers to treatments. Assuming that the logging policy is customer independent, we compare our proposed policy, $\uppi$, with different targeting policies using the following counterfactual policy evaluation metrics: \\
\noindent
(1) Inverse Propensity Score \cite{swaminathan2015counterfactual}: $\displaystyle\frac{1}{N} (\sum_{k=0}^K \displaystyle\frac{\sum_{i=1}^N Z_i\mathbb{1}\{ \pi_{ik} = 1 \text{ \& }  T_i= t_k\}}{p_k})$ \\
\noindent
(2) Self-Normalized IPS \cite{swaminathan2015self}:
$\displaystyle\frac{\sum_{k=0}^K \displaystyle\frac{\sum_{i=1}^N Z_i\mathbb{1}\{ \pi_{ik} = 1 \text{ \& }  T_i= t_k\}}{p_k}}{\sum_{k=0}^K \displaystyle\frac{\sum_{i=1}^N \mathbb{1}\{ \pi_{ik} = 1 \text{ \& }  T_i= t_k\}}{p_k}}$ \\

\noindent
where $Z_i$ can be the outcome of interest (e.g., revenue or completion rate) and $p_k = 1/K, k = 0,1,\ldots, K$.

Depending upon the application, the outcome of interest, $Z_i$, might be different. Typically, it is the same as the objective for uplift models i.e., $Z_i = Y_i$ but sometimes, we may be interested in evaluating more than one business objective. For example, when targeting customers with discount offers, we would want a targeting policy that increases net revenue while maintaining customer experience and in this case, $Z_i$ can reflect revenue or customer experience. We discuss specific business metrics when we study different applications of this problem.

\section{Applications}\label{applications}

We leverage data from large-scale experiments that support different business purposes to test the proposed methodology. We assess the potential impact of given treatments via offline policy estimates. The experiments took the form of an online randomized A/B test, in which the treatment group was given the said treatment and the control group was not. In all experiments, we have binary treatment levels i.e., $\{t_0, t_1\}$ or $\{0,1\}$ for simplicity and estimate the CATE for each customer $i \in \mathcal{C}$ as $\tau(x_i)=\text{E}[Y_i(1) \mid X_i = x_i] - \text{E}[Y_i(0) \mid X_i = x_i]$ using uplift models mentioned in Section~\ref{solution_frame}. All policies are compared based on this randomized experiment data via offline policy evaluation presented in Sec~\ref{eval_metrics}. We discuss the specific details and outcome of each experiment below.

\subsection{Increase Customer Retention}\label{ret_app}

In this scenario, we aim to identify the customers to target with an intervention that improves customer retention. Consider the business case of a company offering paid services aiming to enhance customer engagement and loyalty. The company defines churn as the absence of any service payment within the last month. To mitigate churn, the company intends to identify at-risk customers and engage them proactively with messages promoting additional services. One approach is to predict each customer's likelihood to continue using the services, termed as their retention score. Customers with lower retention scores (specifically below 0.391) are selected for targeted interventions based on this prediction model. While the company has meticulously selected the 0.391 threshold through historical data analysis, relying solely on this retention score threshold for targeting interventions may not be optimal.

\begin{figure}
    \includegraphics[width = 0.5\columnwidth, height = 5cm]{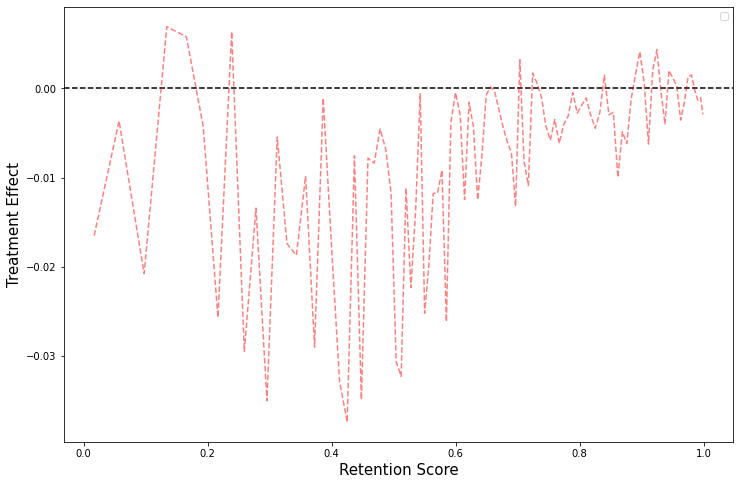}
    \caption{``True'' uplift based on simple difference across different retention scores.}
    \label{fig:ret_dataavg}
\end{figure}

To explore customer reaction to such interventions, we ran a randomized experiment where customers in the treatment group received a messaging template encouraging them to learn more about the services offered by the company. We then clustered the customers into 100 buckets of equal size in an incremental manner going from minimum value of retention scores to maximum and calculated the "true" uplift, $\bar{\tau}^b$, for each  bucket $b$ as 
$\bar{\tau}^b = \frac {\sum_{i \in b} Y_i\mathbb{1}\{T_i=1\}}{\sum_{i \in b} \mathbb{1}\{T_i=1\}} - \frac{\sum_{i \in b} Y_i\mathbb{1}\{T_i=0\}}{\sum_{i \in b} \mathbb{1}\{T_i=0\}}.$ Figure \ref{fig:ret_dataavg} shows this "true" uplift based on simple difference in means along the retention score percentiles. We observe, customers with higher retention scores (> 0.6) have close to 0 uplift, meaning that they are not affected by the targeting message and will probably continue to stay nonetheless. However, the customers with lower scores tend to have negative uplift, which means that the targeting message is doing more harm than good. This shows that the targeting policy based simply on retention score threshold is not optimal for business.

To define optimal targeting segment, we built uplift models using the data from the randomized experiment data. The outcome $Y$ is binary i.e., 1 if the customer does not churn else 0. Intuitively, covariate $X$ for uplift model can be the retention score, as it directly captures the likelihood of the desired outcome. However, sometimes a single variable may not be able to capture the variability in the customers. So, we also explore using the top customer features that were used to build the retention model as the set of covariates, $X$. Since higher uplift means more treatment impact, we aim to target customers with the higher uplift, keeping the proportion of customers to be targeted as low. In this case, the constraint can be to restrict the proportion of customers to be targeted by $c$, if there is a budget constraint. This would essentially mean targeting the top $c \times N$ customers with highest uplift. However, we do not specify $c$ explicitly but rather look for proportion of customers with positive uplift, that can be targeted by different policies. Thus, we do not need a constraint optimization in this case. The proportion of customers we target based on a policy will be referred to as the targeting proportion hereafter. The end goal of targeting customers is to increase retention so we choose the policy with maximum IPS and SNIPS corresponding to $Z_i = Y_i$. Based on uplift model evaluation, we propose a policy based on Causal Forest trained on customer features (see Table~\ref{tab:supp_ret_ope} in Appendix). Just by targeting ~6.5\% of the population, our proposed policy gets a relative lift of 2.35\% in retention rate when compared to the old policy that targets based on retention score, a lift of 3.905\% when compared to not targeting anyone and a lift of 2.79\% in retention rate when compared to targeting everyone (see Table~\ref{tab:ret_ope}).

\begin{table}[!htb]
    \small
    \caption{Relative lift in metrics by the proposed policy over baseline policies for customer retention.}
    \label{tab:ret_ope}
     \begin{minipage}{\columnwidth}
    \begin{tabular}{p{0.2\linewidth}  p{0.12\linewidth} p{0.1\linewidth}}
    \toprule
      \multicolumn{1}{c}{} & \multicolumn{2}{c}{{Lift in Retention Rate}} \\
     \toprule
         {Baseline Policy}  & {IPS} & {SNIPS}  \\
         \toprule
        Retention score < 0.391   & 2.35\% & 0.647\% \\ 
        Targeting no one   & 3.905\% & 1.302\% \\ 
        Targeting everyone   & 2.79\% & 0.215\%  \\
        \bottomrule
    \end{tabular}
    \end{minipage}
\end{table}

\subsection{Maximize Event Revenue}\label{satori_app}

In this business scenario, the objective is to boost customer participation in shopping events and promotional activities. Often times e-commerce websites run deal events, where customers are invited to make purchase over some spend threshold to get certain rewards back on their order. We ran a marketing experiment at a commercial scale, where we offered $P_1$ or $P_2$ rewards back ($P_1 < P_2$) if the customers purchased goods over a fixed amount within a time frame. They randomly assigned $P_1$ or $P_2$ rewards back offer to customers, which wasn't optimal given business goals. Ultimately, the goal was to minimize a population-level efficiency metric defined as, $\text{e\%iS} = \displaystyle\frac{{\text{Average Reward Expense}}}{\text{Average Sale} - \text{Baseline Average Sale}},$ where average sale is attributed to the purchases that happened during the days of the event and baseline average sale refers to the sales had the campaign not run. Since the company had to let everyone participate in the promotion, this baseline average sale was counter-factually calculated. While the goal is to target customer with a suitable offer that improves customer experience and increases revenue, we do not want to deteriorate sales by more than 1\% compared to assigning maximum reward (i.e., $P_2$) for everyone. The customers have a better experience when they participate in these campaigns and are able to complete the offer by purchasing goods worth the defined amount within the allocated time frame and receive the rewards.

Here, control denotes giving $P_1$ and treatment denotes giving $P_2$ rewards. Let $\text{Sales}_i$ be during the deal event and $\text{Rewards}_i$ be rewards for customer $i \in \mathcal{C}$. In order to maximize the overall revenue while controlling the cost, we define outcome $Y$ as $Y_i = \text{Sales}_i - \text{Rewards}_i$. Similar to the application in section 3.1, we train uplift models based on (1) the propensity of a customer to complete the offer (referred as completion score hereafter), and (2) customer features. Assuming each customer is valued the same i.e., $w_i \equiv 1 \ \forall i \in \mathcal{C}$, the optimal policy $\uppi$ can be found such that,
$$\text{max}_\pi \sum_{i=1}^N \pi_{ik} \hat{\tau}(x_i), \text{ subject to } \pi_{ik} \in \{0,1\} \ ,\ \sum_{k=0}^1 \pi_{ik} =1 \ , 1 - \frac{\text{Sales}_\uppi}{\text{Sales}_{\uppi^*}} \leq 0.01,$$
where $\text{Sales}_\uppi = \frac{\sum_{k=0}^1 \sum_{i=1}^N Sale_i\mathbb{1}\{ \pi_{ik} = 1 \text{ \& }  T_i= k\}}{\sum_{k=0}^1 \sum_{i=1}^N \mathbb{1}\{ \pi_{ik} = 1 \text{ \& }  T_i= k\}}$ denotes the average expected sale under the targeting policy $\uppi$ and $\text{Sales}_\uppi^* = \frac{\sum_{i=1}^N Sale_i\mathbb{1}\{T_i= 1\}}{\sum_{i=1}^N \mathbb{1}\{T_i= 1\}}$ denotes the average expected sale when giving all customers the maximum-reward treatment ($P_2$). The constraint here restricts the deterioration of sales by more than 1\% compared to assigning maximum reward for everyone. Since the population is large, the optimization is time consuming. To make it scalable, we create equal-sized groups of customers based on the predicted uplift $\hat{\tau}$ and then optimize over these groups, treating each group as an individual. Consequently, we assign a treatment to a group i.e., everyone in a group receives the same treatment.

We considered a targeting strategy where we can offer higher rewards ($P_2$) to the customers with low propensity to complete the offer so that they're more motivated to shop. Since goal of this campaign is to increase revenue, minimize e\%iS and improve customer experience, we compare our proposed policy with the aforementioned old policy based on these metrics. Based on offline evaluation on the randomized experiment data, we recommend the policy associated with Causal Forest DML trained on completion score (see Table~\ref{tab:supp_satori_ope} in Appendix). Our recommended policy outperforms other policies in terms of revenue and e\%iS, while having a better completion rate than the baseline targeting policy associated with completion scores threshold, see Table~\ref{tab:satori_ope}.

\begin{table}
\small
  \caption{Relative lift in metrics by the proposed policy over baseline policies for revenue maximization.}
    \label{tab:satori_ope}
      \begin{minipage}{\columnwidth}
   \begin{tabular}{p{0.2\linewidth}  p{0.1\linewidth} p{0.1\linewidth} p{0.11\linewidth} p{0.11\linewidth} p{0.12\linewidth}}
    \toprule
      \multicolumn{1}{c}{} & \multicolumn{2}{c}{{Lift in Revenue}} & \multicolumn{2}{c}{{Lift in Completion Rate}} & \multicolumn{1}{c}{}\\
     \toprule
         {Baseline Policy}  & {IPS} & {SNIPS} & {IPS} & {SNIPS} & {e\%iS}   \\
         \toprule
        Completion score < 0.15   & 0.317\% & 0.254\% & 1.02\% & 1.02\%  & -11.879\% \\ 
        Only $P_1$  & 0.993\% & 0.934\% & 2.062\% & 2.062\%  & -55.368\% \\ 
        Only $P_2$   & 0.724\% & 0.665\% & -1.980\% & -1.980\%  & -11.879\% \\
        \bottomrule
    \end{tabular}
    \end{minipage}
\end{table}

\section{Online Experiment: Optimal offer for revenue maximization}\label{online_exp}
Sometimes, customers are offered rewards when they make purchases over a certain amount in shopping events. The objective is to allocate offer optimally to maximize the total revenue while maintaining customer experience. In these shopping events, two thresholds for the purchase amount, say $S_1$ and $S_2$, were randomly assigned to customers where $S_2 > S_1$. An intuitive hypothesis was that lowering the spend threshold will incentivize customers to complete the offer and spend more on average. However, previous campaign data showed that the treatment with $S_1$ spend threshold had $0.30\%$ lower average revenue as compared to $S_2$ spend threshold, while more customers completed the offer with $S_1$. This result motivated us to make optimal offer allocations to customers that maximizes the total revenue from the event, while maintaining customer experience. 

Here, $T=0$ denotes giving spend threshold of $S_2$ and $T=1$ refers to giving spend threshold of $S_1$. Our outcome is $Y$ defined as $Y_i = \text{Sales}_i - \text{Rewards}_i$ and the covariate $X$ can be the completion score or the customer features. We aim to target everyone with a positive uplift and hence, there is no constraint optimization in this case. Based on offline evaluation, we propose a policy using previous campaign data based on Causal Forest trained on customer features (see Table~\ref{tab:supp_fde_ope} in Appendix) because it outperforms all baseline policies in terms of revenue (see Table~\ref{tab:fde_ope}). The proposed policy only assigns the lower spend threshold, $S_1$, to ~12\% of the population while the policy based on completion score assigns it to ~39\% of the population. Note that giving customers $S_1$ purchase threshold, is more costly than offering $S_2$ purchase threshold. Our policy has better completion rate than if we were to offer only $S_2$ threshold and has better revenue than if we were to offer only $S_1$ threshold.

We then conducted a large-scale commercial experiment to assess the effectiveness of our proposed policy. We compared the performance of our recommended targeting policy with a strong static baseline policy of offering $S_2$ spend threshold to all customers based on past campaign (see Table~\ref{tab:supp_past_fde} in Appendix). In this randomized experiment, customers in control (C) group received $S_2$ threshold only, while the treatment (T) group customers were offered either $S_1$ or $S_2$ spend threshold based on our proposed policy. Our proposed policy recommended $S_1$ spend threshold to about 11\% of the customers in the treatment group. The treatment group achieved statistically significant lift in revenue ($0.36\%; p = 0.040$) and completion rate ($5.49\%; p = 0.000$).

\begin{table}
\small
\caption{Relative lift in metrics by the proposed policy when compared with different baseline policies for assigning spend thresholds to customers to receive rewards.}
    \label{tab:fde_ope}
     \begin{minipage}{\columnwidth}
    \begin{tabular}{p{0.25\linewidth}  p{0.12\linewidth} p{0.12\linewidth} p{0.13\linewidth} p{0.13\linewidth}}
    \toprule
      \multicolumn{1}{c}{} & \multicolumn{2}{c}{{Lift in Revenue}} & \multicolumn{2}{c}{{Lift in Completion Rate}}\\
     \toprule
         {Baseline Policy}  & {IPS} & {SNIPS} & {IPS} & {SNIPS} \\
         \toprule
        Completion score < 0.5  & 0.022\% & -0.204\% & -3.017\% & -3.287\%   \\ 
        $S_2$ Only  & 0.214\% & 0.008\% & 2.506\% & 2.221\% \\ 
        $S_1$ Only   & 2.225\% & 2.015\% & -12.494\% & -12.737\% \\
        \bottomrule
    \end{tabular}
    \end{minipage}
\end{table}

\section{Conclusion}\label{summary}

This work provides a principled approach to customer targeting scenarios in e-commerce industry. Based on uplift modeling and constrained optimization, we propose a generalized targeting policy framework for allocating treatments to customers such that the outcome of interest is optimized while taking account of any given business constraints. We demonstrate improvement over baseline targeting approaches in two business applications using offline evaluations and validate our proposed policy via a large-scale online experiment. This framework enables businesses to personalize their marketing campaigns thereby improving value to the business and customer experience.

In all applications, we had binary treatment levels. However, in some cases, we may have infinite continuous treatment levels. For example, one may want to offer customers x\% points, somewhere between say, 5\% to 15\%. There are a few estimation methods, such as double/debiased machine learning \cite{chernozhukov2018double}, that support estimation of continuous treatment effect. With such models, this framework can be naturally extended to the case of continuous treatment levels. Our framework is also applicable to observational studies, given the unconfoundedness assumption, that is, we will need to measure and control for all potential confounders. When the unobserved confounding is a concern, other methods such as proximal causal inference\cite{tchetgen2020introduction} and sensitivity analysis \cite{chernozhukov2022long, zheng2023sensitivity} can be used to make reliable causal conclusions.

\clearpage

\bibliographystyle{ACM-Reference-Format}
\bibliography{acmart}


\begin{thebibliography}{19}


\ifx \showCODEN    \undefined \def \showCODEN     #1{\unskip}     \fi
\ifx \showDOI      \undefined \def \showDOI       #1{#1}\fi
\ifx \showISBNx    \undefined \def \showISBNx     #1{\unskip}     \fi
\ifx \showISBNxiii \undefined \def \showISBNxiii  #1{\unskip}     \fi
\ifx \showISSN     \undefined \def \showISSN      #1{\unskip}     \fi
\ifx \showLCCN     \undefined \def \showLCCN      #1{\unskip}     \fi
\ifx \shownote     \undefined \def \shownote      #1{#1}          \fi
\ifx \showarticletitle \undefined \def \showarticletitle #1{#1}   \fi
\ifx \showURL      \undefined \def \showURL       {\relax}        \fi
\providecommand\bibfield[2]{#2}
\providecommand\bibinfo[2]{#2}
\providecommand\natexlab[1]{#1}
\providecommand\showeprint[2][]{arXiv:#2}

\bibitem[Albert and Goldenberg(2022)]%
        {albert2022commerce}
\bibfield{author}{\bibinfo{person}{Javier Albert} {and} \bibinfo{person}{Dmitri
  Goldenberg}.} \bibinfo{year}{2022}\natexlab{}.
\newblock \showarticletitle{E-commerce promotions personalization via online
  multiple-choice knapsack with uplift modeling}. In
  \bibinfo{booktitle}{\emph{Proceedings of the 31st ACM International
  Conference on Information \& Knowledge Management}}.
  \bibinfo{pages}{2863--2872}.
\newblock


\bibitem[Ascarza(2018)]%
        {ascarza2018retention}
\bibfield{author}{\bibinfo{person}{Eva Ascarza}.}
  \bibinfo{year}{2018}\natexlab{}.
\newblock \showarticletitle{Retention futility: Targeting high-risk customers
  might be ineffective}.
\newblock \bibinfo{journal}{\emph{Journal of Marketing Research}}
  \bibinfo{volume}{55}, \bibinfo{number}{1} (\bibinfo{year}{2018}),
  \bibinfo{pages}{80--98}.
\newblock


\bibitem[Athey et~al\mbox{.}(2019)]%
        {athey2019generalized}
\bibfield{author}{\bibinfo{person}{Susan Athey}, \bibinfo{person}{Julie
  Tibshirani}, {and} \bibinfo{person}{Stefan Wager}.}
  \bibinfo{year}{2019}\natexlab{}.
\newblock \showarticletitle{Generalized random forests}.
\newblock  (\bibinfo{year}{2019}).
\newblock


\bibitem[Chernozhukov et~al\mbox{.}(2018)]%
        {chernozhukov2018double}
\bibfield{author}{\bibinfo{person}{Victor Chernozhukov}, \bibinfo{person}{Denis
  Chetverikov}, \bibinfo{person}{Mert Demirer}, \bibinfo{person}{Esther Duflo},
  \bibinfo{person}{Christian Hansen}, \bibinfo{person}{Whitney Newey}, {and}
  \bibinfo{person}{James Robins}.} \bibinfo{year}{2018}\natexlab{}.
\newblock \bibinfo{title}{Double/debiased machine learning for treatment and
  structural parameters}.
\newblock
\newblock


\bibitem[Chernozhukov et~al\mbox{.}(2022)]%
        {chernozhukov2022long}
\bibfield{author}{\bibinfo{person}{Victor Chernozhukov},
  \bibinfo{person}{Carlos Cinelli}, \bibinfo{person}{Whitney Newey},
  \bibinfo{person}{Amit Sharma}, {and} \bibinfo{person}{Vasilis Syrgkanis}.}
  \bibinfo{year}{2022}\natexlab{}.
\newblock \bibinfo{booktitle}{\emph{Long story short: Omitted variable bias in
  causal machine learning}}.
\newblock \bibinfo{type}{{T}echnical {R}eport}. \bibinfo{institution}{National
  Bureau of Economic Research}.
\newblock


\bibitem[Devriendt et~al\mbox{.}(2021)]%
        {devriendt2021you}
\bibfield{author}{\bibinfo{person}{Floris Devriendt}, \bibinfo{person}{Jeroen
  Berrevoets}, {and} \bibinfo{person}{Wouter Verbeke}.}
  \bibinfo{year}{2021}\natexlab{}.
\newblock \showarticletitle{Why you should stop predicting customer churn and
  start using uplift models}.
\newblock \bibinfo{journal}{\emph{Information Sciences}}  \bibinfo{volume}{548}
  (\bibinfo{year}{2021}), \bibinfo{pages}{497--515}.
\newblock


\bibitem[Du et~al\mbox{.}(2019)]%
        {du2019improve}
\bibfield{author}{\bibinfo{person}{Shuyang Du}, \bibinfo{person}{James Lee},
  {and} \bibinfo{person}{Farzin Ghaffarizadeh}.}
  \bibinfo{year}{2019}\natexlab{}.
\newblock \showarticletitle{Improve User Retention with Causal Learning}. In
  \bibinfo{booktitle}{\emph{The 2019 ACM SIGKDD Workshop on Causal Discovery}}.
  PMLR, \bibinfo{pages}{34--49}.
\newblock


\bibitem[Goldenberg et~al\mbox{.}(2020)]%
        {goldenberg2020free}
\bibfield{author}{\bibinfo{person}{Dmitri Goldenberg}, \bibinfo{person}{Javier
  Albert}, \bibinfo{person}{Lucas Bernardi}, {and} \bibinfo{person}{Pablo
  Estevez}.} \bibinfo{year}{2020}\natexlab{}.
\newblock \showarticletitle{Free lunch! retrospective uplift modeling for
  dynamic promotions recommendation within roi constraints}. In
  \bibinfo{booktitle}{\emph{Proceedings of the 14th ACM Conference on
  Recommender Systems}}. \bibinfo{pages}{486--491}.
\newblock


\bibitem[Gutierrez and G{\'e}rardy(2017)]%
        {gutierrez2017causal}
\bibfield{author}{\bibinfo{person}{Pierre Gutierrez} {and}
  \bibinfo{person}{Jean-Yves G{\'e}rardy}.} \bibinfo{year}{2017}\natexlab{}.
\newblock \showarticletitle{Causal inference and uplift modelling: A review of
  the literature}. In \bibinfo{booktitle}{\emph{International conference on
  predictive applications and APIs}}. PMLR, \bibinfo{pages}{1--13}.
\newblock


\bibitem[Oprescu et~al\mbox{.}(2019)]%
        {oprescu2019orthogonal}
\bibfield{author}{\bibinfo{person}{Miruna Oprescu}, \bibinfo{person}{Vasilis
  Syrgkanis}, {and} \bibinfo{person}{Zhiwei~Steven Wu}.}
  \bibinfo{year}{2019}\natexlab{}.
\newblock \showarticletitle{Orthogonal random forest for causal inference}. In
  \bibinfo{booktitle}{\emph{International Conference on Machine Learning}}.
  PMLR, \bibinfo{pages}{4932--4941}.
\newblock


\bibitem[Rubin(2005)]%
        {rubincausal2005}
\bibfield{author}{\bibinfo{person}{Donald~B Rubin}.}
  \bibinfo{year}{2005}\natexlab{}.
\newblock \showarticletitle{Causal Inference Using Potential Outcomes}.
\newblock \bibinfo{journal}{\emph{J. Amer. Statist. Assoc.}}
  \bibinfo{volume}{100}, \bibinfo{number}{469} (\bibinfo{year}{2005}),
  \bibinfo{pages}{322--331}.
\newblock


\bibitem[Rzepakowski and Jaroszewicz(2012)]%
        {rzepakowski2012decision}
\bibfield{author}{\bibinfo{person}{Piotr Rzepakowski} {and}
  \bibinfo{person}{Szymon Jaroszewicz}.} \bibinfo{year}{2012}\natexlab{}.
\newblock \showarticletitle{Decision trees for uplift modeling with single and
  multiple treatments}.
\newblock \bibinfo{journal}{\emph{Knowledge and Information Systems}}
  \bibinfo{volume}{32} (\bibinfo{year}{2012}), \bibinfo{pages}{303--327}.
\newblock


\bibitem[Swaminathan and Joachims(2015a)]%
        {swaminathan2015counterfactual}
\bibfield{author}{\bibinfo{person}{Adith Swaminathan} {and}
  \bibinfo{person}{Thorsten Joachims}.} \bibinfo{year}{2015}\natexlab{a}.
\newblock \showarticletitle{Counterfactual risk minimization: Learning from
  logged bandit feedback}. In \bibinfo{booktitle}{\emph{International
  Conference on Machine Learning}}. PMLR, \bibinfo{pages}{814--823}.
\newblock


\bibitem[Swaminathan and Joachims(2015b)]%
        {swaminathan2015self}
\bibfield{author}{\bibinfo{person}{Adith Swaminathan} {and}
  \bibinfo{person}{Thorsten Joachims}.} \bibinfo{year}{2015}\natexlab{b}.
\newblock \showarticletitle{The self-normalized estimator for counterfactual
  learning}.
\newblock \bibinfo{journal}{\emph{advances in neural information processing
  systems}}  \bibinfo{volume}{28} (\bibinfo{year}{2015}).
\newblock


\bibitem[Tchetgen et~al\mbox{.}(2020)]%
        {tchetgen2020introduction}
\bibfield{author}{\bibinfo{person}{Eric J~Tchetgen Tchetgen},
  \bibinfo{person}{Andrew Ying}, \bibinfo{person}{Yifan Cui},
  \bibinfo{person}{Xu Shi}, {and} \bibinfo{person}{Wang Miao}.}
  \bibinfo{year}{2020}\natexlab{}.
\newblock \showarticletitle{An introduction to proximal causal learning}.
\newblock \bibinfo{journal}{\emph{arXiv preprint arXiv:2009.10982}}
  (\bibinfo{year}{2020}).
\newblock


\bibitem[Wager and Athey(2018)]%
        {wager2018estimation}
\bibfield{author}{\bibinfo{person}{Stefan Wager} {and} \bibinfo{person}{Susan
  Athey}.} \bibinfo{year}{2018}\natexlab{}.
\newblock \showarticletitle{Estimation and inference of heterogeneous treatment
  effects using random forests}.
\newblock \bibinfo{journal}{\emph{J. Amer. Statist. Assoc.}}
  \bibinfo{volume}{113}, \bibinfo{number}{523} (\bibinfo{year}{2018}),
  \bibinfo{pages}{1228--1242}.
\newblock


\bibitem[Yadav et~al\mbox{.}(2021)]%
        {yadav2021optimal}
\bibfield{author}{\bibinfo{person}{Amulya Yadav}, \bibinfo{person}{Roopali
  Singh}, \bibinfo{person}{Nikolas Siapoutis}, \bibinfo{person}{Anamika
  Barman-Adhikari}, {and} \bibinfo{person}{Yu Liang}.}
  \bibinfo{year}{2021}\natexlab{}.
\newblock \showarticletitle{Optimal and non-discriminative rehabilitation
  program design for opioid addiction among homeless youth}. In
  \bibinfo{booktitle}{\emph{Proceedings of the Twenty-Ninth International
  Conference on International Joint Conferences on Artificial Intelligence}}.
  \bibinfo{pages}{4389--4395}.
\newblock


\bibitem[Zhao and Harinen(2019)]%
        {zhao2019uplift}
\bibfield{author}{\bibinfo{person}{Zhenyu Zhao} {and} \bibinfo{person}{Totte
  Harinen}.} \bibinfo{year}{2019}\natexlab{}.
\newblock \showarticletitle{Uplift modeling for multiple treatments with cost
  optimization}. In \bibinfo{booktitle}{\emph{2019 IEEE International
  Conference on Data Science and Advanced Analytics (DSAA)}}. IEEE,
  \bibinfo{pages}{422--431}.
\newblock


\bibitem[Zheng et~al\mbox{.}(2023)]%
        {zheng2023sensitivity}
\bibfield{author}{\bibinfo{person}{Jiajing Zheng}, \bibinfo{person}{Jiaxi Wu},
  \bibinfo{person}{Alexander D’Amour}, {and} \bibinfo{person}{Alexander
  Franks}.} \bibinfo{year}{2023}\natexlab{}.
\newblock \showarticletitle{Sensitivity to Unobserved Confounding in Studies
  with Factor-structured Outcomes}.
\newblock \bibinfo{journal}{\emph{J. Amer. Statist. Assoc.}}
  (\bibinfo{year}{2023}), \bibinfo{pages}{1--12}.
\newblock


\end{thebibliography}

\appendix
\renewcommand\thefigure{\thesection.\arabic{figure}}
\renewcommand\thetable{\thesection.\arabic{table}}

\section{Appendix}

\subsection{Offline Application Results}

\begin{table}[H]
    \caption{The targeting proportion and offline policy evaluation metrics for the different policies for member retention.}
    \label{tab:supp_ret_ope}
     \begin{minipage}{\columnwidth}
    \begin{tabular}{p{0.2\linewidth} p{0.2\linewidth} p{0.12\linewidth} p{0.1\linewidth} p{0.1\linewidth}}
    \toprule
      \multicolumn{3}{c}{} & \multicolumn{2}{c}{Retention Rate} \\
     \toprule
         Policy & Targeting proportion & AUC & IPS & SNIPS  \\
         \midrule
      \multicolumn{5}{c}{Based on retention scores} \\
         \midrule
        FDR Learner & 4.22\% & 0.00231 & 93.2\% & 93.2\% \\
        Causal Forest & 6.29\% & 0.00226 & 93.7\% & 93.2\% \\
        Causal Forest DML & 34.69\% & 0.00200 & 92.4\% & 93.0\% \\
        \midrule
        \multicolumn{5}{c}{Based on customer features} \\
        \midrule
        FDR Learner & 0.48\% & 0.00344 & 93.3\% & 93.2\% \\
        \textbf{Causal Forest} & \textbf{6.47\%} & \textbf{0.00355} & \textbf{95.8\%} & \textbf{93.4\%} \\
        Causal Forest DML & 30.57\% & 0.00335 & 92.6\% & 93.1\% \\
        \bottomrule
    \end{tabular}
    \end{minipage}
\end{table}

\begin{table}[H]
    \caption{The targeting proportion, AUC and offline policy evaluation metrics for the different policies for the deal with different points back offer.}
    \label{tab:supp_satori_ope}
    \begin{minipage}{\columnwidth}
    \begin{tabular}{p{0.2\linewidth}  p{0.2\linewidth} p{0.1\linewidth} p{0.1\linewidth} p{0.1\linewidth}  p{0.1\linewidth}}
    \toprule
      \multicolumn{3}{c}{} & \multicolumn{2}{c}{Revenue}  & \multicolumn{1}{c}{}\\
     \toprule
         Policy & Targeting proportion & AUC & IPS & SNIPS  & e\%iS   \\
         \midrule
      \multicolumn{6}{c}{Based on completion scores} \\
         \midrule
        FDR Learner & 8.83\% & 2.371 &  1402.91 & 1402.42   & 88.35\%\\
        Causal Forest & 94.48\% & 0.938 &  1411.67 & 1410.52  & 48.41\%\\
        \textbf{Causal Forest DML} & 67.47\% & 4.373 &  \textbf{1417.27} & \textbf{1416.44}  & \textbf{35.46\%} \\
        \midrule
        \multicolumn{6}{c}{Based on customer features} \\
        \midrule
        FDR Learner & 97.85\% & -0.023 &  1412.02 & 1412.17   & 45.20\%\\
        Causal Forest & 10.91\% & \textbf{4.670} &  1408.70 & 1409.03   & 52.22\%\\
        Causal Forest DML & 99.49\% & -0.673 &  1412.35 & 1412.01   & 45.30\%\\
        \bottomrule
    \end{tabular}
    \end{minipage}
\end{table}

\begin{table}[H]
    \caption{The targeting proportion, AUC and offline policy evaluation metrics for the different policies for the deal with different spend thresholds.}
    \label{tab:supp_fde_ope}
    \begin{minipage}{\columnwidth}
    \begin{tabular}{p{0.2\linewidth}  p{0.2\linewidth} p{0.1\linewidth} p{0.1\linewidth} p{0.1\linewidth} }
    \toprule
      \multicolumn{3}{c}{} & \multicolumn{2}{c}{Revenue}  \\
     \toprule
         Policy & Targeting proportion & AUC & IPS & SNIPS    \\
         \midrule
      \multicolumn{5}{c}{Based on completion scores} \\
         \midrule
        FDR Learner & 34.49\% & 5.748 & 4,646.53 & 4,643.30 \\
        Causal Forest & 28.75\% & 29.455 & 4,686.02 & 4,678.98 \\
        Causal Forest DML & 33.85\% & 10.144 & 4,647.33 & 4,646.82 \\
        \midrule
        \multicolumn{5}{c}{Based on customer features} \\
        \midrule
        FDR Learner & 12.79\% & 21.788 & 4,680.94 & 4,662.42 \\
        \textbf{Causal Forest} & 11.93\% & \textbf{35.804} & \textbf{4,698.25} & \textbf{4,688.58} \\
        Causal Forest DML & 11.70\% & 22.566 & 4,676.99 & 4,646.82 \\
        \bottomrule
    \end{tabular}
    \end{minipage}
\end{table}

\subsection{Online Experiment Results}

\begin{table}[H]
\caption{Past experiment results comparing static spend threshold of $S_2$ (C) vs $S_1$ (T).}
\label{tab:supp_past_fde}
 \begin{minipage}{\columnwidth}
\begin{tabular}{p{0.2\linewidth}  p{0.16\linewidth} p{0.2\linewidth}}
\toprule
  & Revenue & Completion Rate \\ \midrule
Lift (T-C)     & -0.25\%   & 18.18\%  \\ 
p-value (T-C)     & 0.098   & 0.000   \\ \bottomrule
\end{tabular}
\end{minipage}
\end{table}


We deep-dive further into how our proposed policy might have performed if we were to assign every one, whether in control or treatment groups, the policy recommended threshold of $S_1$ or $S_2$ (see Table \ref{tab:model-deepdive}). The completion rate of customers who are recommended $S_2$ threshold, is not very different ($2.75\%$ uplift) for control and treatment groups. However, the completion rate of customers who are recommended $S_1$ threshold, is very different ($18.37\%$ uplift) for control and treatment groups. This could be attributed to the fact that these customers were much more likely to complete the offer when given the $S_1$ spend threshold.

\begin{table}[H]
\caption{Difference in completion rates for customers who the model recommended $S_1$ vs $S_2$ spend threshold in control and treatment groups}
\label{tab:model-deepdive}
\begin{minipage}{\columnwidth}
\begin{tabular}{p{0.16\linewidth}  p{0.3\linewidth}}
\toprule
Threshold & Completion Rate Uplift (T-C) \\ \midrule
$S_1$  & 18.37\% \\
$S_2$  & 2.74\%  \\ \bottomrule
\end{tabular}
\end{minipage}
\end{table}


\end{document}